\documentclass{article}

\usepackage{microtype}
\usepackage{graphicx}
\usepackage{tabularx}
\usepackage{subcaption}
\usepackage{booktabs} 
\usepackage{microtype}
\usepackage{graphicx}
\usepackage{booktabs} 
\usepackage{amsmath,amssymb}
\usepackage{url}
\usepackage{color}
\usepackage[noend]{algpseudocode}
\usepackage{todonotes}
\usepackage{enumitem}

\usepackage{hyperref}
\usepackage{cleveref}

\newcommand{\tvi}{TVI}

\usepackage[accepted]{icml2019}

\icmltitlerunning{Learning Robot Skills with Temporal Variational Inference}

\begin{document}

\twocolumn[
\icmltitle{Learning Robot Skills with Temporal Variational Inference}




\begin{icmlauthorlist}
\icmlauthor{Tanmay Shankar}{fb}
\icmlauthor{Abhinav Gupta}{fb,cmu}
\end{icmlauthorlist}

\icmlaffiliation{fb}{Facebook AI Research, Pittsburgh, PA, USA}
\icmlaffiliation{cmu}{Carnegie Mellon University, Pittsburgh, PA, USA}

\icmlcorrespondingauthor{Tanmay Shankar}{tanmayshankar@fb.com}

\icmlkeywords{Machine Learning, ICML}

\vskip 0.3in
]

\printAffiliationsAndNotice{} 

\begin{abstract}
In this paper, we address the discovery of robotic options from demonstrations in an unsupervised manner.
Specifically, we present a framework to jointly learn low-level control policies and higher-level policies of how to use them from demonstrations of a robot performing various tasks. 
By representing options as continuous latent variables, we frame the problem of learning these options as latent variable inference. We then present a temporal formulation of variational inference based on a temporal factorization of trajectory likelihoods, that allows us to infer options in an unsupervised manner. 
We demonstrate the ability of our framework to learn such options across three robotic demonstration datasets. 
\end{abstract}
\section{Introduction}
The robotics community has long since sought to acquire general purpose and reusable robotic skills, to enable robots to execute a wide range of tasks.
The idea of such skills is attractive; by abstracting away the details of low-level control and reasoning over high-level skills instead, a robot can address more complex and longer term tasks. 
Further, 
the ability to \textit{compose} skills leads to a combinatorial increase in the robot's capabilities \cite{tlpkICRA17}, ideally spanning the abilities required for the robot to complete the desired tasks.  
For example, a robot equipped with reaching, grasping, and pouring skills could compose them to make a cup of tea as well as pour cereal into a bowl.

Indeed, the promise of skills has been explored in several contexts,
be it options in reinforcement learning (RL) \cite{sutton1999between}, operators in the planning \cite{fikes1971strips}, primitives and behaviours in  robotics \cite{5686298}, or abstractions \cite{NIPS2019_9332}. Nomenclature aside, these ideas share the notion of eliciting a certain pattern or template of action in response to a particular situation, differing in their exact implementations and how these skills are obtained. 
For example, while previous works \cite{5686298, doi:10.1177/0278364912472380, konidaris2009skill} manually defined these skills, more recent approaches have  sought to \textit{learn} these skills, either from interaction \cite{hrl_tejas} or from demonstrations \cite{xu2018neural, huang2019neural, fox2017multi, krishnan2017ddco, kipf2019compile}. 

Learning skills from demonstrations is appealing, since it allows non-robotics-expert humans to demonstrate solving the target tasks, bypassing the tedium of manually specifying these skills or carefully engineering solutions to the tasks \cite{argall2009survey}. But even using demonstrations, approaches such as \citet{xu2018neural, huang2019neural} require heavy supervision such as segmentation of demonstration data. 
Instead, learning skills in an unsupervised, data-driven manner not only avoids having to annotate demonstrations; it also enables the use of large scale and diverse demonstration data in robotics \cite{pmlr-v87-sharma18a, mandlekar2018roboturk}. This in turn enables learning a correpsondingly diverse set of skills, as well as how to use them to achieve a variety of tasks. 

Aside from the issue of learning and representing individual skills, is the notion of \textit{composing} them. The efficacy of these skills is rather limited when used in isolation; by selecting and sequencing the appropriate skills however, a robot can achieve a variety of complex tasks, as illustrated by the tea and cereal example. 
A rich body of literature addresses composing skills, including sequencing skills \cite{neumann2014learning, peters2013towards}, hierarchical approaches \cite{doi:10.1177/0278364917713116, xu2018neural, huang2019neural, konidaris2012robot, konidaris2009skill, shankar2020discovering}, options \cite{sutton1999between}, etc. 
Some works among these \cite{fox2017multi, krishnan2017ddco, kipf2019compile} address jointly learning skills and how to compose them. Jointly learning both levels of this hierarchy not only addresses how to use these skills, but also allows for adapting the skills based on how useful they are to the task at hand. 

Unfortunately, these works have their own limitations. 
While \citet{neumann2014learning, niekum2012learning, konidaris2009skill} do learn to sequence skills, they assume restrictive primitive representations - such as DMPs \cite{ijspeert2013dynamical}. While \citet{shankar2020discovering} learn continuous representations of primitives, it requires an additional post hoc policy learning step. \citet{fox2017multi, krishnan2017ddco} do afford directly usable policies, but critically are restricted to a fixed number of discrete options. 

In this paper, we propose a framework to jointly learn options and how to compose and use them from demonstrations in an unsupervised manner. At the heart of our framework is a \textit{temporal variational inference} (\tvi) based on a corresponding factorization of trajectory likelihoods. 
We adopt a latent variable representation of options; this allows us to treat the problem of inferring options as inferring latent variables, which we may then accomplish via our temporal variational inference.
Specifically, optimizing the objective afforded by our temporal variational inference with respect to the distributions involved naturally gives us low and high-level policies of options. 

We evaluate our approach's ability to learn options across three datasets, and demonstrate that our approach can learn a meaningful space of options that correspond with traditional skills in manipulation, visualized at 
\href{https://sites.google.com/view/learning-causal-skills/home}{https://sites.google.com/view/learning-causal-skills/home}. 
We quantify the effectiveness of our policies in solving downstream tasks, evaluated on a suite of tasks. 

\section{Related Work}
\textbf{Learning from Demonstrations:}
Learning from Demonstrations (LfD) addresses solving tasks by learning from demonstrations of the task being solved by an expert. This may be accomplished by simply cloning the original demonstration \cite{esmaili1995behavioural}, or fitting the demonstration to a trajectory representation \cite{kober2009learning, peters2013towards} or a policy \cite{Schaal_ANIPS_1997}. 
More recent efforts have sought to segment demonstrations into smaller behaviors, and fitting a model to the resulting segments \cite{niekum2012learning, krishnan2018transition, murali2016tsc, meier2011movement}. \citet{argall2009survey} presents a thorough review of these techniques. Our work falls into the broad paradigm of LfD, but seeks to learn hierarchical policies from demonstrations. 

\textbf{Sequencing Primitives:}
The concept of learning movement primitives using predefined representations of primitives \cite{kober2009learning, peters2013towards} has been a popular technique to capturing robotic behaviors. 
A natural next step is to sequence such primitives to perform longer horizon downstream tasks, \cite{neumann2014learning, niekum2012learning, 5686298}. \citet{konidaris2009skill, konidaris2012robot} address merging these skills into skill-trees. 
\citet{doi:10.1177/0278364917713116, rudolf2} address sequencing primitives using probabilistic segmentation and attribute grammars respectively. 
Our work differs from a majority of these works in that we jointly learn both the representation of primitives and how these primitives must be sequenced. 

\textbf{Hierarchical Policy Learning:}
\citet{fox2017multi, krishnan2017ddco, kipf2019compile, DBLP:conf/iclr/SharmaSRK19} also address the problem of learning options from demonstrations without supervision. However these works are restricted to a discrete set of options, which must be pre-specified. Further, \citet{fox2017multi, krishnan2017ddco} employ a forward-backward algorithm for inference that requires an often intractable marginalization over latents. The CompILE framework \cite{kipf2019compile} makes use of continuous latent variables to parameterize options, but requires carefully designed attention mechanisms, and is evaluated in relatively low-dimensional domains. 
\citet{smith2018inference} and \citet{bacon2017option} derive policy gradients to address hierarchical policy learning in the RL setting.

\textbf{Learning Trajectory Representations:}
\citet{shankar2020discovering} learning representations of primitives, but need to adopt an additional phase of training to produce usable policies. \citet{co2018self} also approach hierarchical RL from a trajectory representation perspective. Our work shares this notion of implicitly learning a representation of primitives, but differs in that we do not require an additional high-policy learning step to perform hierarchical control. 

\textbf{Learning Temporal Abstractions:}
\citet{NIPS2019_9332} and \citet{gregor2018temporal} both address learning temporal abstractions in the form of `jumpy' transitions. \citet{NIPS2019_9332} seeks to learn a generic partitioning of the input sequence into temporal abstractions, while \citet{gregor2018temporal} learns temporal abstractions with beliefs over state to capture uncertainity about the world. While both these works adopt variational bounds of a similar form to ours, our bound objective is derived in terms of usable option \textit{policies} rather than abstract or belief states. 

\textbf{Compositional Policies:} 
Both \citet{xu2018neural} and \citet{huang2019neural} address unseen manipulation tasks by learning compositional policies, but require heavy supervision to do so. 
\citet{andreas2017modular} and \citet{shiarlis2018taco} compose modular policies in the RL and LfD settings respectively, using policy sketches to select which policies to execute.
While our work approaches doesn't explicitly address compositionality, we too seek to benefit from the benefits of such compositionality. 
\section{Approach}


\subsection{Preliminaries}
Throughout our paper, we adopt an undiscounted Markov Decision Process without rewards (denoted as MDP\textbackslash R). An MDP\textbackslash R  is a tuple $\mathcal{M}: \langle S, A, P \rangle$, that consists of states $s$ in state space $S$, actions $a$ in action space $A$, and a transition function between successive states $P(s_{t+1} | s_t, a_t)$.
 
\renewcommand{\algorithmicrequire}{\textbf{Input:}}
\renewcommand{\algorithmicensure}{\textbf{Output:}}
\algnewcommand{\LineComment}[1]{\Statex \(\triangleright\) #1}
\algnewcommand{\LineComment2}[1]{\Statex \hspace*{0.45\linewidth} \(\triangleright\) #1}
\algnewcommand{\LineComment3}[1]{\Statex \hspace*{\fill} \(\triangleright\) #1}

\begin{algorithm}[t]
	\caption{Trajectory Generation Process with Options}
	\label{alg:Alg1}
	\begin{algorithmic}[1]	
	    \Require Low-level Policy $\pi$, High-level Policy $\eta$, Initial State Distribution $d_1(s)$, 
	    \Ensure Trajectory $\tau$
        
        \State $s_1 \sim d_1$, $b_1 \gets 1$ \Comment{Initialize state.}
		\For {$t \in [1,2,..,T]$}
	    \If {$b_t = 1$}:
	        \State $z_t \sim \eta(z|s_{1:t}, a_{1:t-1}, z_{1:t-1})$ \Comment{Select option.}
	    \Else: 
	        \State $z_t \gets z_{t-1}$ \Comment{Continue previous option.}
	    \EndIf
	    \State $a_t \sim \pi(a|s_{1:t}, a_{1:t-1}, z_{1:t})$ \Comment{Select action.}
	    \State $s_{t+1} \sim p(s_{t+1} | s_{t}, a_{t})$ \Comment{Execute action.}
	    \State $b_{t+1} \sim \beta(s_{t+1})$ \Comment{Decide whether to terminate.}
		\EndFor
	    \State $\tau \gets \{ s_t, a_t \}_{t=1}^T$, $\zeta \gets \{z_t, b_t\}_{t=1}^T$
	\end{algorithmic}
\end{algorithm}

\textbf{Options:} An option $\omega \in \Omega$ \cite{sutton1999between} formally consist of three components - an initiation set $\mathcal{I}$, a policy $\pi: S \rightarrow A$, and a termination function $\beta: S \rightarrow [0,1]$. When an option is invoked in a state in $\mathcal{I}$, policy $\pi$ is executed until the termination function dictates the option should be terminated (i.e., $\beta(s)=1$). As in \citet{fox2017multi, smith2018inference}, we assume options may be initiated in the entire state space. 

\textbf{Options as Latent Variables:} 
We assume that the identity of an option being executed is specified by a latent variable $z$, that may be either continuous or discrete. 
We also consider that at every timestep, a high-level policy $\eta: S \rightarrow \Omega \times [0,1]$ selects the identity $z_t$ of the option to be invoked, as well as a binary variable $b_t$ of whether to terminate the option or not. This option informs the low-level policy $\pi$'s choice of action, constructing a trajectory as per the generative process in \cref{alg:Alg1}.
We denote the sequence of options executed during a trajectory as a sequence of these latent variables, $\zeta = \{z_t, b_t\}_{t=1}^T$.


\begin{figure*}[t!]
\begin{center}
\centerline{\includegraphics[width=\linewidth]{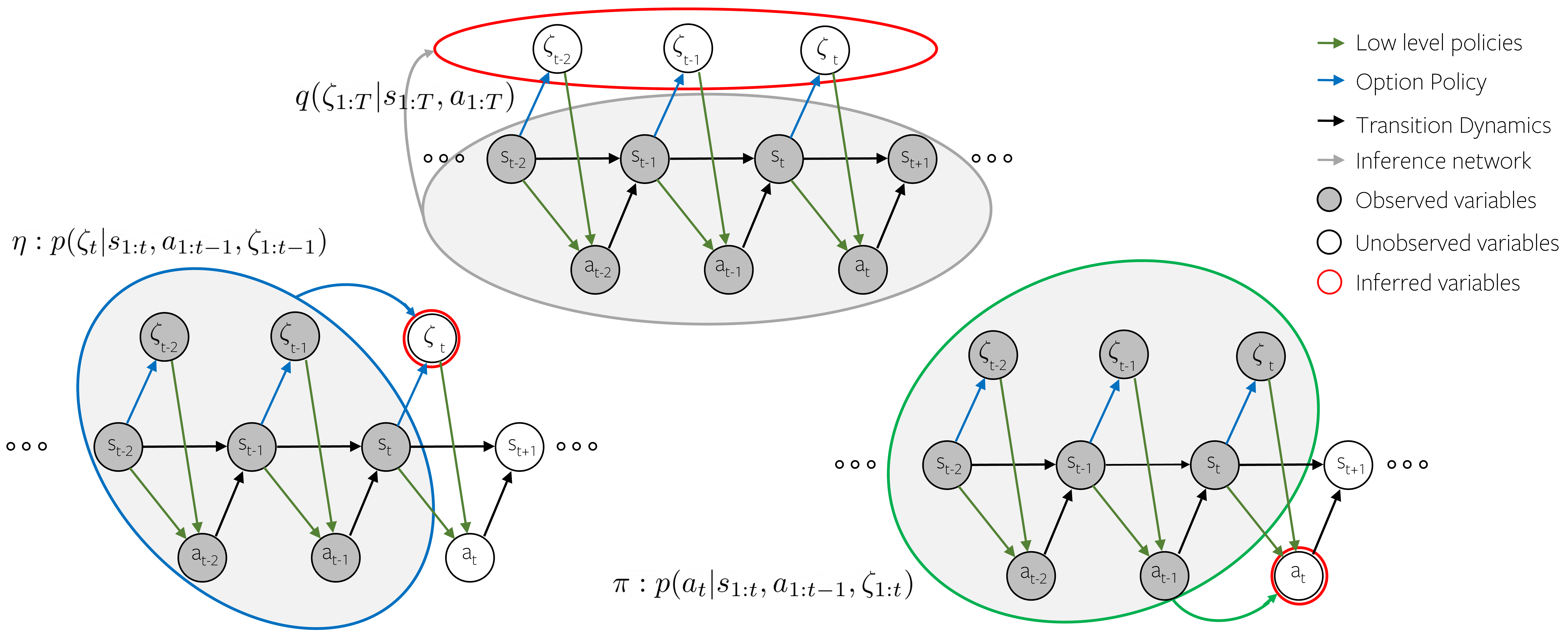}}
\caption{Depiction of the key distributions $q$, $\pi$, and $\eta$, and the probabilistic graphical model underlying our approach. Note the dependence between trajectories and options. We also depict the variables that each of the three networks reason about, and the information they make use of to do so.}
\label{causal-pgm}
\end{center}
 \vskip -0.2in
\end{figure*}



\subsection{Decomposition of trajectory likelihood}
\label{sec:causal_decomp}
Consider a trajectory $\tau = \{ s_t, a_t \}_{t=1}^T$, and the sequence option sequence that generated it $\zeta = \{ z_t, b_t\}_{t=1}^T$. The joint likelihood $p(\tau, \zeta)$ of the trajectory and these options under the generative process described in \cref{alg:Alg1} may be expressed as follows: 
\begin{equation}
\begin{split}
    p(\tau, \zeta) = p(s_1) \prod_{t=1}^{T} \eta(\zeta_t | s_{1:t}, a_{1:t-1}, \zeta_{1:t-1}) \\
    \pi(a_t | s_{1:t}, a_{1:t-1}, \zeta_{1:t}) p(s_{t+1} | s_t, a_t) 
    \label{eq:traj_likelihood}
\end{split}
\end{equation}

The distributions $\eta(\zeta_t | s_{1:t}, a_{1:t-1}, \zeta_{1:t-1})$ and $\pi(a_t | s_{1:t}, a_{1:t-1}, \zeta_{1:t})$ implicitly capture the causal restriction that future variables (ex. $s_{t+1:T}$) do not influence earlier variables (ex. $a_{1:t}$). Further, both $\pi$ and $\eta$ may be queried at any arbitrary time $t$ with the information available till that time. 
\citet{6479340} formalized the notion of \textit{causally conditioned distributions} to represent such distributions, 
a notion that plays a part in the formulation of our approach. We refer the reader to \citet{6479340, Kramer1998DirectedIF} for a more thorough treatment of this concept. 


\subsection{Temporal Variational Inference}
To reiterate, our goal is to learn options and how to use them from demonstrations; this formally corresponds to learning low and high level policies $\pi$ and $\eta$, from a dataset of $N$ demonstrations, $\mathcal{D} = \{\tau_i\}_{i=1}^N$.
This is equivalent to inferring latent variables $\zeta$ from a trajectory, since policies $\pi$ and $\eta$ essentially reason about the choice of $\zeta$ and its effect on the choice of actions. 
The representation of options as latent variables $\zeta$ we adopt hence allows us to view the problem of inferring options from a perspective of inference of latent variables. 

This allows us to employ unsupervised latent variable inference techniques; we employ a variant of variational inference (VI) \cite{kingma2013autoencoding} to infer latent options $\zeta$. 
Our choice of VI over the forward-backward style algorithm employed in \citet{fox2017multi, krishnan2017ddco} is because VI is amenable to both continuous and discrete latent variables $z$. Further, VI bypasses the often intractable marginalization over latents in favor of sampling based approach. 

In standard VI, a variational distribution $q(z|x)$ is used to infer latents $z$ from observed data $x$, approximating the unknown conditional $p(z|x)$. One then optimizes the likelihood of observations given the predicted latents under a learned decoder $p(x|z)$. 
In our case, we seek to infer the \textit{sequence} of options $\zeta = \{z_t,b_t\}_{t=1}^T$ from a trajectory $\tau = \{s_t,a_t\}_{t=1}^T$; we estimate the conditional $p(\zeta | \tau)$ with a variational approximation $q(\zeta | \tau)$.


To retrieve usable policies that can be queried at inference or ``test" time,
we require policies that can reason about the current choices of option $\zeta_t$ and action $a_t$ given the observations available \textit{so far}, i.e. $s_{1:t}$, $a_{1:t-1}$, and $\zeta_{1:t-1}$. 
This precludes optimizing the conditional $p(\tau|\zeta)$ as would be done in standard VI. Instead, we optimize the joint likelihood $p(\tau, \zeta)$ of trajectories and latents with respect to the causally conditioned $\pi$ and $\eta$. 
Not only does this afford us directly usable policies as desired, but this objective naturally arises from the variational bound constructed below.


We now formally present temporal variational inference, a variant of VI suitable for our sequential latent variables, that accounts for the causal restriction of these latent variables based on the decomposition of trajectory likelihood in \cref{sec:causal_decomp}. 
We begin with the standard objective of maximizing the log-likelihood of trajectories across the dataset, $\mathcal{L} = \mathbb{E}_{\tau \sim \mathcal{D}} \big[ \log p(\tau) \big]$.
$\mathcal{L}$ is lower bounded by $J$, where:
\begin{align*}
    J &= \mathbb{E}_{\tau \sim \mathcal{D}} \big[ \log p(\tau) \big]  - D_{KL} \big[ q(\zeta|\tau) || p(\zeta|\tau) \big] \\
    &= \mathbb{E}_{\tau \sim \mathcal{D}} \big[ \log p(\tau) \big]  - \mathbb{E}_{\tau \sim \mathcal{D}, \zeta \sim q(\zeta| \tau)} \Big[ \log \frac{q(\zeta|\tau)}{p(\zeta|\tau)} \Big] \\
    &= \mathbb{E}_{\tau \sim \mathcal{D}, \zeta \sim q(\zeta| \tau)} \Big[ \log p(\tau) + \log p(\zeta | \tau) - \log q(\zeta|\tau) \Big] \\
    &= \mathbb{E}_{\tau \sim \mathcal{D}, \zeta \sim q(\zeta| \tau)} \Big[ \log p(\tau, \zeta) - \log q(\zeta|\tau) \Big]
\end{align*}
Substituting the joint likelihood decomposition from \cref{eq:traj_likelihood} above yields the following objective: 
\begin{align}
    J &= \mathbb{E}_{\tau \sim \mathcal{D}, \zeta \sim q(\zeta|\tau)} \Big[ \sum_t \{ \log \eta(\zeta_t | s_{1:t}, a_{1:t-1}, \zeta_{1:t-1}) \nonumber \\ 
    &+ \log \pi(a_t | s_{1:t}, a_{1:t-1}, \zeta_{1:t}) + \log p(s_{t+1} | s_t, a_t) \} \nonumber \\
    &+ \log p(s_1) - \log q(\zeta|\tau) \Big]
    \label{eq:objective}
\end{align}
Assuming distributions $\pi, \eta$, and $q$ are parameterized by $\theta, \phi$, and $\omega$ respectively, we may optimize these using standard gradient based optimization of $J$:
\begin{align}
    \nabla J &= \nabla_{\theta, \phi, \omega} \mathbb{E}_{\tau \sim \mathcal{D}, \zeta \sim q(\zeta|\tau)} \Big[ \sum_t \{ \log \pi(a_t | s_{1:t}, a_{1:t-1}, \zeta_{1:t}) \nonumber \\ 
    &+ \log \eta(b_t, z_t | s_{1:t}, a_{1:t-1}, \zeta_{1:t-1}) \} - \log q(\zeta|\tau) \Big] \ \ \ 
    \label{eq:gradient}
\end{align}
Note that the dynamics $p(s_{t+1}|s_t,a_t)$ and initial state distribution $p(s_1)$ factor out of this gradient, as derived in the supplementary material.

\textbf{Parsing the objective $J$:}
We provide a brief analysis of how objective $J$ and its implied gradient update \cref{eq:gradient} jointly optimizes $\pi$, $\eta$, and $q$ to be consistent with each other. 
Three interacting terms optimize $q$. The first two terms encourage $q$ to
predict options $\zeta$ that result in high likelihood of actions $a_t$ under $\pi$, and that are likely under the current estimate of the high level policy $\eta$. The final $-\log q(\zeta|\tau)$ term encourages maximum entropy of $q$, discouraging $q$ from committing to an option unless it results in high likelihood under $\pi$ and $\eta$. This entropy term also prevents $q$ from trivially encoding all trajectories into a single option. 

Low-level policy $\pi$ is trained to increase the likelihood of selecting actions $a_t$ given the current option being executed $\zeta_t$. High-level policy $\eta$ is trained to mimic the choices of options made by the variational network 
(i.e. options that result in high likelihood of demonstrated actions), only using the available information at time $t$ to do so. 

\textbf{Reparameterization:}
While \cref{eq:gradient} can be implemented via REINFORCE \cite{williams1992simple}, we do so only for inferring discrete variables $\{b_t\}_{t=1}^T$. 
As in standard VI, we exploit the continuous nature of $z$'s and employ the reparameterization trick \cite{kingma2013autoencoding} to enable efficient learning of $\{z_t\}_{t=1}^T$. 
Rather than sample latents $\{z_t\}_{t=1}^T$ from a stochastic variational distribution $q(\zeta|\tau)$, $\{z_t\}_{t=1}^T$ is parameterized as a differentiable and deterministic function of the inputs $\tau$ and a noise vector $\epsilon$, drawn from an appropriately scaled normal distribution.
In practice, we parameterize $q$ as an LSTM that takes $\tau$ as input, and predicts mean $\mu_t$ and variance $\sigma_t$ of the distribution 
$q(\{z_t\}_{t=1}^T|\tau)$. We then retrieve $\{z_t\}_{t=1}^T$ as $\{ z_t = \mu_t + \sigma_t \epsilon_t \}_{t=1}^T$.

In our case, gradients flow from our objective $J$ through \textit{both}
the low and high-level policies $\pi$ and $\eta$ to the variational network $q$. This in contrast with standard VI, where gradients pass through the decoder $p(x|z)$. 
This reparameterization enables the efficient gradient based learning of $q$ based on signal from both the low and high-level policies.

\textbf{Features of objective:}
The temporal variational inference we present has several desirable traits that we describe below. 
\newline
(1) First and foremost, $J$ provides us with causally conditioned low and high-level policies $\pi$ and $\eta$. These policies are directly usable at inference time towards solving downstream tasks, since they are only trained with information that is also available at inference time. 
\newline
(2) With \tvi, we can adopt a continuous parameterization of options, $z \in \mathbb{R}^n$. This eliminates the need to pre-specify the number of options required; instead we may learn as many options as are required to capture the behaviors observed in the data, in a data driven manner. The continuous space of options also allows us to reason about how similar the various learned options are to one another (allowing substitution of options for one another). 
\newline
(3) The joint training of $\pi$, $\eta$, and $q$ implied by $J$ not only allows training the high-level policy $\eta$ to based on the available options, but also allows the adaptation of the low-level options $\pi$ based on how useful they are to reconstructing a demonstration. 
\newline
(4) The objective $J$ is also amenable to gradient based optimization, allowing us to learn options efficiently. 



\begin{algorithm}[t]
	\caption{Temporal Variational Inference for Learning Skills}
	\label{alg:alg2}
	\begin{algorithmic}[1]	
	    \Require $\mathcal D$ \Comment{Require a demonstration dataset}
	    \Ensure $\pi, \eta$ \Comment{Output low and high-level policies}

        
        \State Initialize $\pi_{\theta}, \eta_{\phi}, q_{\omega}$ \Comment{Initialize networks}
        \State Pretrain $\pi$ as VAE \Comment{Pretrain latent representation}
		\For {$i \in [1,2,...,N_{\rm iterations}]$}
		\State $\tau_i \gets \mathcal{D}$ \Comment{Retrieve trajectory from dataset}
		\State $\zeta \sim q(\zeta|\tau_i)$ \Comment{Sample latent sequence from variational network}
        \State $J \gets \sum_t \log \pi(a_t|s_{1:t},a_{1:t-1},\zeta_{1:t}) + \sum_t \log \eta (\zeta_t|s_{1:t},a_{1:t-1},\zeta_{1:t-1})  - \log q(\zeta|\tau)$ 
        \LineComment3{Evaluate likelihood objective under current policy estimates}
        \State Update $\pi_{\theta}, \eta_{\phi}, q_{\omega}$ via $\nabla_{\theta, \phi, \omega} J$ 
		\EndFor
	\end{algorithmic}
\end{algorithm}



\subsection{Learning Skills with Temporal VI}
Equipped with this understanding of our temporal variational inference (\tvi), we may retrieve low and high-level policies $\pi$ and $\eta$ by gradient based optimization of the objective $J$ presented. We first make note of some practical considerations required to learn skills with \tvi, then present a complete algorithm to learn skills using \tvi.

\textbf{Policy Parameterization:}
We parameterize each of the policies $\pi$ and $\eta$ as LSTMs \cite{lstm}, with $8$ layers and $128$ hidden units per layer. The recurrent nature of the LSTM naturally captures the causal nature of $\pi$ and $\eta$. In contrast, $q$ is parameterized as a bi-directional LSTM, since $q$ reasons about the sequence of latents $\zeta$ given the entire trajectory $\tau$. 
$q$, $\pi$ and $\eta$ all take in a concatenation of trajectory states and actions as input. $\pi$ and $\eta$ also take in the sequence of latents until the current timestep as input. Since $\eta$ reasons about the choice of latents to solve the task occurring in the demonstration, $\eta$ also takes in additional information about the task, such as task ID and object-state information. 
$\pi$ predicts the mean $\mu_a$ and variance $\sigma_a$ of a Gaussian distribution from which actions $a \in \mathcal{A}$ are drawn. $\eta$ and $q$ both predict mean $\mu_z$ and variance $\sigma_z$ of a Gaussian distribution from which latent variables $z$ are drawn. $\eta$ and $q$ also predict the probability of terminating a particular option $p(b)$, from which binary termination variables $b$ are drawn. While $q$ predicts the entire sequence of $\zeta$'s, latents are retrieved from $\eta$ during a rollout via the generative process in \cref{alg:Alg1}.

\textbf{Pretraining the Low-level Policy:} 
Optimizing our joint objective $J$ with randomly initialized low-level policies results in our training procedure diverging. During initial phases of training, the random likelihoods of actions and options under the random initial policies provide uninformative gradients. 
To counteract this, we initialize the low-level policy to capture some meaningful skills, by pretraining it to reconstruct demonstration segments in a VAE setting. Specifically, we draw trajectory segments from the dataset and encode them as a single latent $z$ (i.e. a single option). We train the low-level policy (i.e. as a decoder) to maximize the likelihood of the actions observed in this trajectory segment given the latent $z$ predicted by the encoder.

\begin{figure*}[t!]
    \centering
    \begin{subfigure}[t]{0.32\textwidth}
        \centering
        \includegraphics[height=\linewidth]{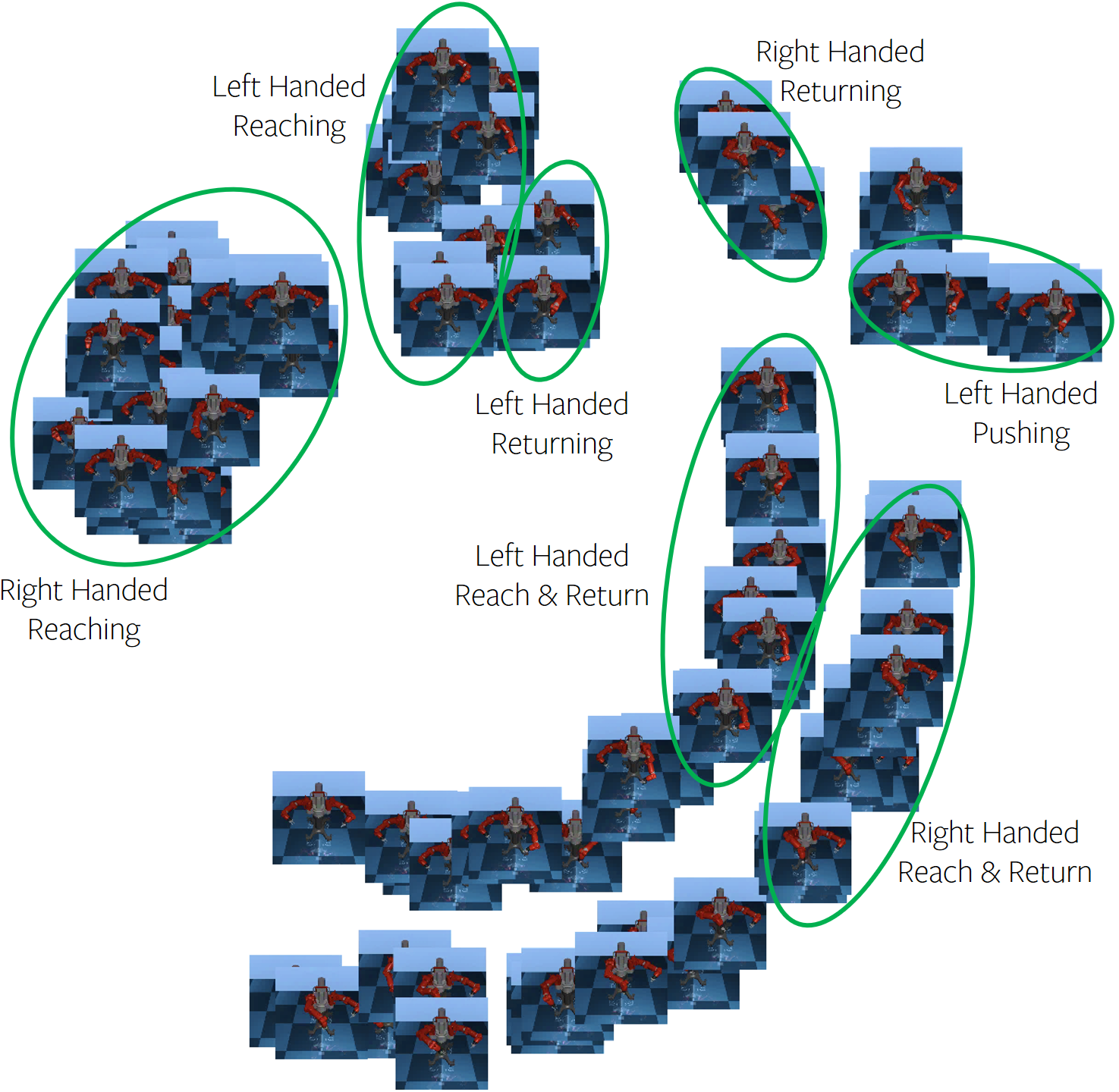}
        \caption{Latent space of skills for MIME dataset. Note the clustering of skills into left and right handed reaching, returning and sliding skills, along with additional hybrid skills.}
        \label{fig:MIMEEmbedding}
    \end{subfigure}%
    \hfill
    \begin{subfigure}[t]{0.32\textwidth}
        \centering
        \includegraphics[width=\linewidth]{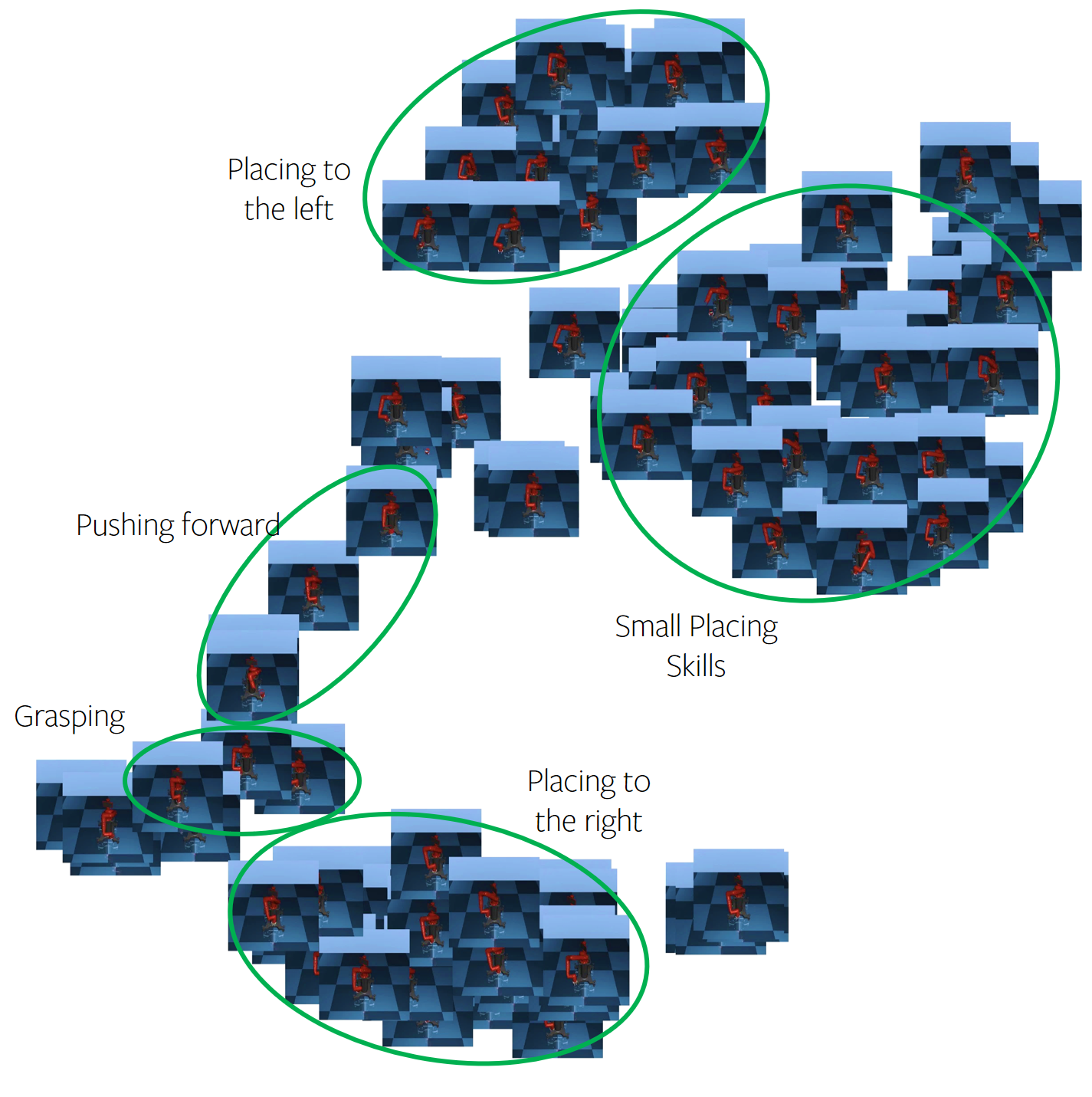}
        \caption{Latent space of skills for Roboturk dataset. Note the clustering of skills into single armed pushing, grasping, and placing in different locations and to different extents.}
        \label{fig:RTurkEmbedding}
    \end{subfigure}%
    \hfill
    \begin{subfigure}[t]{0.32\textwidth}
        \centering
        \includegraphics[width=\linewidth]{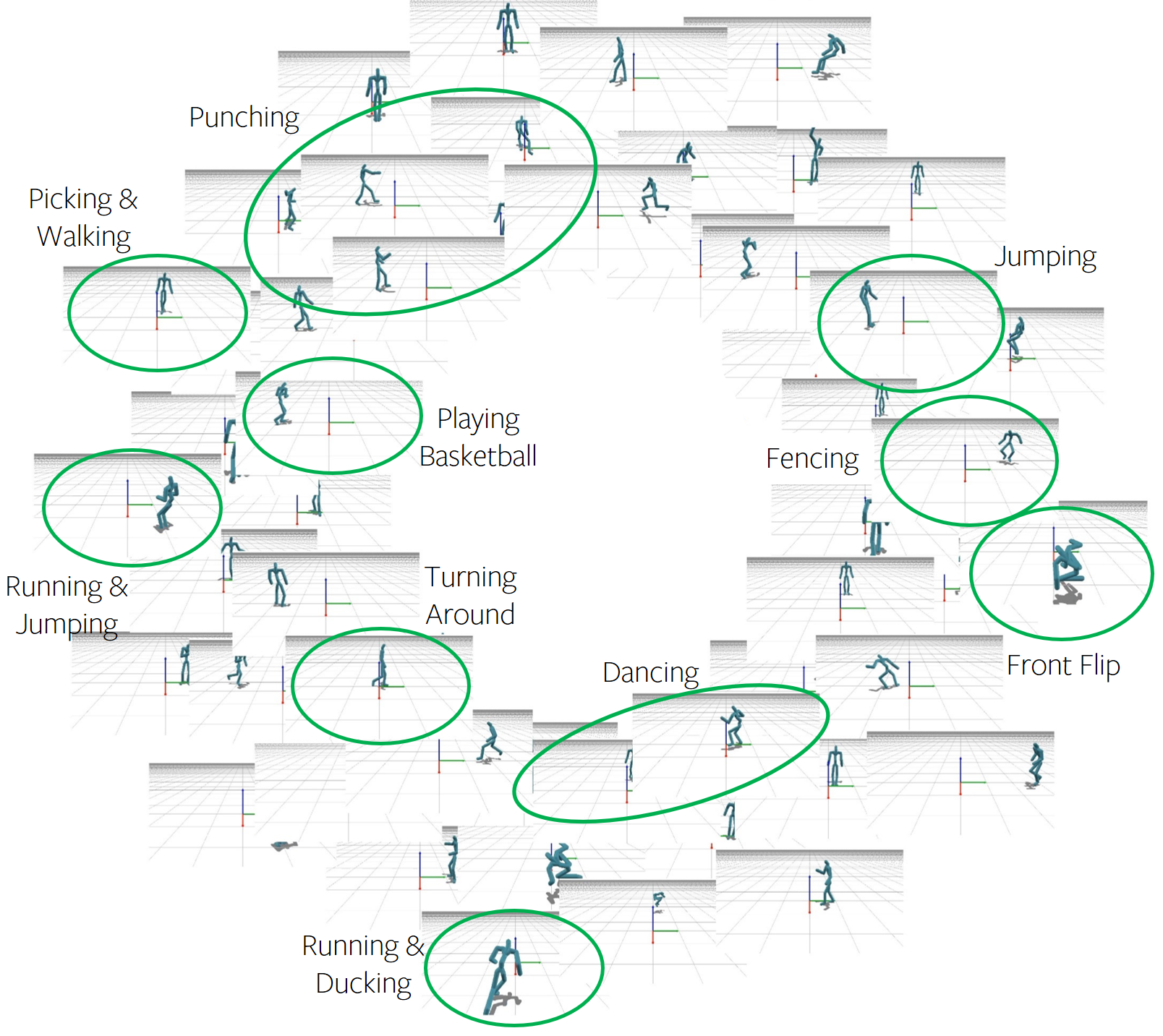}
        \caption{Latent space of skills for Mocap dataset. While less structured than (a) and (b), the space consists of diverse skills ranging such as running, front flips and punching.}
        \label{fig:MocapEmbedding}
    \end{subfigure}%
    \caption{Visualization of the learned latent space of skills for the (a) MIME dataset, (b) Roboturk dataset, and (c) Mocap dataset. Note the emergence of clusters of skills in each case.}
\end{figure*}

\textbf{Algorithm:} We present the full algorithm for learning skills via temporal variational inference in \cref{alg:alg2}. After the pre-training step described above, there are three steps that occur every iteration in training. (1) For every trajectory, a corresponding sequence of latent variables $\zeta$ is sampled from the varitional network $q$. (2) This likelihood of this estimated sequence of latents $\zeta$ is evaluated under the current policy estimates, giving us objective $J$. (3) The gradients of $J$ are then used to update the three networks $\pi$, $\eta$, and $q$.

\section{Experiments}
We would like to understand how well our approach can discover options from a set of demonstrations in an unsupervised manner, and quantify how useful the learnt policies are for solving a set of target tasks. To this end, we evaluate our approach across the three datasets described below, as well as a suite of simulated robotic tasks. We present visualizations of the results of our model at \href{https://sites.google.com/view/learning-causal-skills}{https://sites.google.com/view/learning-causal-skills}. We first describe the datasets used below.
\newline
\textbf{MIME Dataset:} The MIME Dataset \cite{pmlr-v87-sharma18a} consists of 8000+ kinesthetic demonstrations across $20$ tasks (such as pushing, bottle-opening, stacking, etc.) collected on a Baxter robot. We use the $16$ dimensional joint-angles of the Baxter ($7$ joints for each of the $2$ arms, and a gripper for each arm) as the input and prediction space for our model.
\newline
\textbf{Roboturk Dataset:} The Roboturk Dataset \cite{mandlekar2018roboturk} consists of 2000+ demonstrations collected on a suite of tasks (such as bin picking, nut-and-peg assembly, etc.) by teleoperating a simulated Sawyer robot. These teleoperation demonstrations are hence more noisy than the kinesthetic MIME dataset. We use the $8$ dimensional joint angles of the Sawyer ($7$ arm joints, and a single gripper), as well as the robot-state and object-state vectors provided in \citet{mandlekar2018roboturk} to train our model.
\newline
\textbf{CMU Mocap Dataset:} The CMU Mocap Dataset \cite{cmu_mocap} consists of 1953 motions collected by tracking visual markers placed on humans, while performing a variety of different actions. These actions include punching, jumping, performing flips, running, etc. We use the local (i.e., relative to a root node on the agent itself) 3-D positions of each of the $22$ joints recorded in the dataset, for a total of $66$ dimensions. 
\newline
\textbf{Preprocessing and Train-Test Split:}
In both the MIME and Roboturk datasets, the gripper values are normalized to a range of $\{-1,1\}$, while joint angles are unnormalized. The trajectories across all datasets are downsampled (in time) by a factor of $20$. 
For each dataset, we set aside $500$ randomly sampled trajectories that serve as our test set for our experiments in \cref{subsec:Recon}. The remaining trajectories serve as the respective training sets.


\subsection{Qualitative Evaluation of Learned Space}
The first question we would like to answer is - ``Is our approach able to learn a diverse space of options?". We answer this by presenting a qualitative analysis of the space of options learned by our model. 

For a set of $100$ demonstrations, we retrieve the sequence of latent $z$'s and their corresponding trajectory segments executed over each demonstration from our model. We embed these latent $z$'s in a 2-D space using T-SNE \cite{maaten2008visualizing}, and visualize the trajectory segments at their corresponding position in this 2-D embedded space.
We visualize the learned embedding spaces for the MIME dataset in \cref{fig:MIMEEmbedding}, for the Roboturk dataset in \cref{fig:RTurkEmbedding}, and for the Mocap dataset in \cref{fig:MocapEmbedding}. Dynamic visualizations of these figures are available at \href{https://sites.google.com/view/learning-causal-skills}{https://sites.google.com/view/learning-causal-skills}.

Note the emergence of \textit{clusters} of skills on the basis of types of motions being executed across these datasets. 

In the case of the MIME dataset \cref{fig:MIMEEmbedding}, we note that the emergent skills are separated on the basis of the nature of motion being executed, as well as the \textit{arm} of the robot being used. This is expected in a bimanual robot, as skills with the left and right hands are to be treated differently, as are skills using both hands. 
The skills that our approach captures correspond directly to traditional notions of skills in the manipulation community, such as reaching, returning, sliding / pushing, etc. Our approach further distinguishes between left handed reaching and right handed reaching, etc., a useful ability in addressing downstream tasks. 

For the Roboturk dataset \cref{fig:RTurkEmbedding}, the space is separated on the basis of the nature, shape and direction of motion being executed. For example, placing motions to the left and right of the robot appear separately in the space. Additional placing motions that move the arm to a lesser extent also appear separately. The space also captures finer motions such as closing the grippers down (typically in a position above the workspace). 

\begin{table}[t!]
\caption{Trajectory Reconstruction Error of our approach and baselines across various datasets. The baselines were adapted from (1) \citet{kingma2013autoencoding}, (2) \citet{ijspeert2013dynamical}, (3) \citet{niekum2012learning}, (4) \citet{shankar2020discovering}.}
\label{tab:reconstruction}
\vskip 0.15in
\begin{center}
\begin{small}
\begin{sc}
\begin{tabular}{lcccr}
\toprule
Method & MIME & Roboturk & Mocap  \\
       & Data & Data     & Data \\
\midrule
Flat VAE \hspace{0.01in} [1]    & 0.14      & 0.23  &  0.08 \\
Flat DMP \hspace{0.01in} [2]    & 0.36      & 0.54  &  3.45 \\
H-DMP    \hspace{0.04in} [3]    & 0.02      & 0.06  &  0.01 \\
Disco-MP \hspace{0.01in} [4]    & 0.02      & 0.03  &  0.03 \\
\textbf{Ours}                   & 0.02      & 0.04  &  0.04 \\

\bottomrule
\end{tabular}
\end{sc}
\end{small}
\end{center}
\vskip -0.1in
\end{table}

For the Mocap dataset \cref{fig:MocapEmbedding}, the learned space is not as clearly structured as in the case of MIME and Roboturk datasets. We believe this is due to the much larger range of motions executed in the dataset, coupled with the high-dimensional nature of the humanoid agent. Despite this lack of structure, a diverse set of skills is still learned. For example, the space consists of running, walking, and jumping skills, which constitute a majority of the dataset. More interesting skills such as fencing, performing flips, and boxing skills were also present and captured well by our model.

In both the MIME and Roboturk datasets, the correspondence of many emergent skills with traditional notions of skills in the manipulation community is indicative that our approach can indeed discover diverse robotic skills from demonstrations without supervision. 


\begin{table*}[t]
\caption{Average Rewards of our approach and baselines on various RL environments, across 100 episodes. Baselines are adapted from (1) \citet{lillicrap2015continuous}, (2) \citet{esmaili1995behavioural}, (3) \citet{hrl_tejas}.}
\label{tab:RL}
\vskip 0.15in
\begin{center}
\begin{small}
\begin{tabular}{lcccccc}
\toprule
Method & SawyerPick- & SawyerPick- &  SawyerPick- & SawyerPick- & SawyerNut- & SawyerNut- \\
       & PlaceBread & PlaceCan &  PlaceCereal & PlaceMilk & AssemblyRound & AssemblySquare \\
\midrule
Flat RL [1]         & 0.11 & 0.34 & 0.18 & 0.14 & 0.44 & 0.55 \\
Flat IL [2]         & 0.60 & 0.65 & 0.29 & 0.67 & 0.49 & 1.87 \\
Hierarchical RL (No Init) [3] & 0.13 & 0.11 & 0.04 & 0.15 & 0.36 & 1.16 \\
Hierarchical RL (W/ Init) [3] & 0.41 & 0.37 & 0.22 & 0.34 & 0.54 & 1.09 \\
\textbf{Ours}       & 1.54 & 1.22 & 1.54 & 0.48 & 1.88 & 3.62 \\
\bottomrule
\end{tabular}
\end{small}
\end{center}
\vskip -0.1in
\end{table*}

\subsection{Reconstruction of Demonstrations}
\label{subsec:Recon}
We evaluate how well our approach can use the learned skills to reconstruct the demonstrated trajectories; i.e. whether it is able to capture the overall structure of the demonstrations in terms of the skills being executed. 
We do so quantitatively and qualitatively. Quantitatively, we measure the average state distance (measured by the mean squared error) between the reconstructed trajectory and the original demonstration, across the $500$ randomly sampled unseen demonstrations in each dataset. We compare our approach's performance on this metric against a set of baselines:

\begin{itemize}[noitemsep,topsep=0pt]
    \item \textbf{Flat-VAE:} We train an LSTM VAE \cite{kingma2013autoencoding} to reconstruct demonstrations. This represents a flat, non-hierarchical baseline with a learned representation. The architecture used is an $8$ layer LSTM with $128$ hidden units, like our policies. 
    \item \textbf{Flat-DMP:} We fit a single Dynamic Movement Primitive \cite{ijspeert2013dynamical} to each trajectory in the dataset. This represents a non-hierarchical baseline with a predefined trajectory representation. 
    \item \textbf{H-DMP:} We evaluate a hierarchical DMP based approach similar to \citet{niekum2012learning}, that first segments a trajectory by detecting changepoints in acceleration, and then fits individual DMPs to \textit{each} segment of the trajectory. 
    This represents a hierarchical baseline with a predefined trajectory representation. 
    \item \textbf{Disco-MP:} We also evaluate the method of  \citet{shankar2020discovering}, which is directly optimized for trajectory reconstruction. This approach represents a hierarchical baseline with a learned trajectory representation. 
    \item \textbf{Ours:} We obtain predicted trajectories from our model obtaining the latent z's that occur in a demonstration from our $q$ network, and rolling out the low-level policy with these z's as input.
\end{itemize}

We present the average state distances obtained by our approach against these baselines in \cref{tab:reconstruction}. 
The flat VAE is able to achieve a reasonably low reconstruction error, indicating the benefits of a learnt trajectory representation over a predefined representation such as DMPs.
Combinining a learnt trajectory representation with the ability to compose primitives naturally leads to a further decrease in the trajectory reconstruction error, as observed in the Disco-MP baseline and our approach.
Note that our approach is able to achieve a similar reconstruction error to that of Disco-MP, which is explicitly optimized to minimize (aligned) state distances, as well as the H-DMP baseline, which heavily overfits to a single trajectory (thus promising low reconstruction error). This demonstrates that our approach indeed captures the overall structure of demonstrations and is able to represent them faithfully. These trends are consistently observed across all three datasets we evaluate on.

To qualitatively analyse how well our approach captures the structure of demonstration, we visualize the reconstructed trajectories predicted by our approach against the corresponding ground truth trajectory. These results are presented in our website:
\href{https://sites.google.com/view/learning-causal-skills/home}{https://sites.google.com/view/learning-causal-skills/home}.
We observe that our approach is indeed able to capture the rough sequence of skills that occur in the demonstrated trajectories. In the case of the MIME and Roboturk datasets, our model predicts similar reaching, returning, sliding etc. primitives when the ground truth trajectory executes a corresponding primitives. Further, the rough shape of the arms during the predicted skills correlate strongly with the shapes observed in the ground truth trajectories; this is consistent with the quantitative results presented above. Our approach also notably captures fine motions (such as opening and closing of the gripper) in trajectories well. 
In case of the Mocap dataset, the overall shape and trend of rolled out trajectories align very closely with the original demonstration, showing 
the use of our approach in learning skills across widely differently structured data and morphology of agents. 


\subsection{Downstream Tasks}
We would also like to evaluate how useful our learned policies are for solving a set of target tasks. This is central to our motivation of jointly learning options \textit{and} how to use them. 
We test our learned policies on a suite of $6$ simulated robotic tasks from the Robosuite \cite{mandlekar2018roboturk} environment on the Sawyer robot. 
Since the MIME dataset lacks any object information, we disregard the Baxter tasks in Robosuite \cite{mandlekar2018roboturk}, and instead use tasks from Robosuite for which the Roboturk dataset has demonstrations. 
In the pick-place tasks, a reward of $1$ is given to objects successfully placed in the bin. In the nut-and-peg assembly tasks, a reward of $1$ is given for successfully placing a nut. Both tasks also have additional rewards based on conditions of how the task was executed. We point the reader to \cite{mandlekar2018roboturk} for a full description of these tasks and their reward structure. 
We compare the performance of our method on these tasks against the following baselines: 
\begin{itemize}[noitemsep, topsep=0pt]
    \item \textbf{Flat RL:} We train flat policies on each of the $6$ tasks in the  reinforcement learning setting, using DDPG \cite{lillicrap2015continuous}.
    \item \textbf{Flat IL:} 
    We train flat policies to mimic the actions observed in the demonstrations of each of the $6$ tasks, and subsequently finetune these policies in the RL setting. 
    \item \textbf{Hierarchical RL:} We train hierarchical policies as in \cite{hrl_tejas} in the pure RL setting, with (W/ Init) and without (No Init) any prior initialization.
    \item \textbf{Ours:} We fine-tune the low and high-level policies obtained by our model in the RL setting.
\end{itemize}
All baseline policies are implemented as 8 layer LSTMs with 128 hidden units, for direct comparison with our policies. The RL based approaches are trained with DDPG with 
the same exploration processes and hyperparameters (such as initializations of the networks, learning rates used, etc.), as noted in the supplementary. 
We evaluate each of these baselines along with our approach by testing out the learned policies over $100$ episodes in the $6$ environments, and reporting the average rewards obtained in \cref{tab:RL}. 

We observe that the Flat RL and Hierarchical RL baselines are unable to solve these tasks, and achieve low rewards across all tasks. This is unsurprising given the difficulty of exploration problem underlying these tasks \cite{mandlekar2018roboturk}. 
Pretraining policies with imitation learning somewhat allieviates this problem, as observed in the slightly higher rewards of the Flat IL baseline. This is likely because the policies are biased towards actions similar to those seen in the demonstrations. 
Training hierarchical policies initialized with our approach is able to achieve significantly higher rewards than both the IL and RL baselines consistently across most environments. 
By providing these policies with suitable notions of skills that extend over several timesteps, we are able to bypass reasoning over low-level actions for 100's of timesteps, thus guiding exploration in these tasks more efficiently.

\section{Conclusion}
In this paper, we presented a framework for jointly learning robotic skills and how to use them from demonstrations in an unsupervised manner. Our temporal variational inference allows us to construct an objective that directly affords us usable policies on optimization. We are able to learn semantically meaningful skills that correspond closely with the traditional notions of skills observed in manipulation. Further, our approach is able to capture the overall structure of demonstrations in terms of the learned skills. We hope that these factors contribute towards accelerating research in robot learning for manipulation. 

\section*{Acknowledgements}
The authors would like to thank Shubham Tulsiani for valuable discussions on the formulation of the approach, and Jungdam Won and Deepak Gopinath for help with data processing and visualization for the Mocap dataset. 

\bibliography{References}

\begin{thebibliography}{47}
\providecommand{\natexlab}[1]{#1}
\providecommand{\url}[1]{\texttt{#1}}
\expandafter\ifx\csname urlstyle\endcsname\relax
  \providecommand{\doi}[1]{doi: #1}\else
  \providecommand{\doi}{doi: \begingroup \urlstyle{rm}\Url}\fi

\bibitem[Andreas et~al.(2017)Andreas, Klein, and Levine]{andreas2017modular}
Andreas, J., Klein, D., and Levine, S.
\newblock Modular multitask reinforcement learning with policy sketches.
\newblock In \emph{ICML}, 2017.

\bibitem[Argall et~al.(2009)Argall, Chernova, Veloso, and
  Browning]{argall2009survey}
Argall, B.~D., Chernova, S., Veloso, M., and Browning, B.
\newblock A survey of robot learning from demonstration.
\newblock \emph{Robotics and autonomous systems}, 2009.

\bibitem[Bacon et~al.(2017)Bacon, Harb, and Precup]{bacon2017option}
Bacon, P.-L., Harb, J., and Precup, D.
\newblock The option-critic architecture.
\newblock In \emph{AAAI}, 2017.

\bibitem[CMU(2002)]{cmu_mocap}
CMU.
\newblock Cmu graphics lab motion capture database.
\newblock 2002.
\newblock URL \url{http://mocap.cs.cmu.edu}.

\bibitem[Co-Reyes et~al.(2018)Co-Reyes, Liu, Gupta, Eysenbach, Abbeel, and
  Levine]{co2018self}
Co-Reyes, J.~D., Liu, Y., Gupta, A., Eysenbach, B., Abbeel, P., and Levine, S.
\newblock Self-consistent trajectory autoencoder: Hierarchical reinforcement
  learning with trajectory embeddings.
\newblock \emph{arXiv preprint arXiv:1806.02813}, 2018.

\bibitem[Esmaili et~al.(1995)Esmaili, Sammut, and
  Shirazi]{esmaili1995behavioural}
Esmaili, N., Sammut, C., and Shirazi, G.
\newblock Behavioural cloning in control of a dynamic system.
\newblock IEEE, 1995.

\bibitem[Fikes \& Nilsson(1971)Fikes and Nilsson]{fikes1971strips}
Fikes, R.~E. and Nilsson, N.~J.
\newblock Strips: A new approach to the application of theorem proving to
  problem solving.
\newblock \emph{Artificial intelligence}, 1971.

\bibitem[Fox et~al.(2017)Fox, Krishnan, Stoica, and Goldberg]{fox2017multi}
Fox, R., Krishnan, S., Stoica, I., and Goldberg, K.
\newblock Multi-level discovery of deep options.
\newblock \emph{arXiv preprint arXiv:1703.08294}, 2017.

\bibitem[Gregor et~al.(2019)Gregor, Papamakarios, Besse, Buesing, and
  Weber]{gregor2018temporal}
Gregor, K., Papamakarios, G., Besse, F., Buesing, L., and Weber, T.
\newblock Temporal difference variational auto-encoder.
\newblock In \emph{International Conference on Learning Representations}, 2019.
\newblock URL \url{https://openreview.net/forum?id=S1x4ghC9tQ}.

\bibitem[Higgins et~al.()Higgins, Matthey, Pal, Burgess, Glorot, Botvinick,
  Mohamed, and Lerchner]{higgins2017beta}
Higgins, I., Matthey, L., Pal, A., Burgess, C., Glorot, X., Botvinick, M.,
  Mohamed, S., and Lerchner, A.
\newblock beta-vae: Learning basic visual concepts with a constrained
  variational framework.

\bibitem[Hochreiter \& Schmidhuber(1997)Hochreiter and Schmidhuber]{lstm}
Hochreiter, S. and Schmidhuber, J.
\newblock Long short-term memory.
\newblock \emph{Neural Comput.}, 9\penalty0 (8):\penalty0 1735–1780, November
  1997.
\newblock ISSN 0899-7667.
\newblock \doi{10.1162/neco.1997.9.8.1735}.
\newblock URL \url{https://doi.org/10.1162/neco.1997.9.8.1735}.

\bibitem[Huang et~al.(2019)Huang, Nair, Xu, Zhu, Garg, Fei-Fei, Savarese, and
  Niebles]{huang2019neural}
Huang, D.-A., Nair, S., Xu, D., Zhu, Y., Garg, A., Fei-Fei, L., Savarese, S.,
  and Niebles, J.~C.
\newblock Neural task graphs: Generalizing to unseen tasks from a single video
  demonstration.
\newblock In \emph{CVPR}, 2019.

\bibitem[Ijspeert et~al.(2013)Ijspeert, Nakanishi, Hoffmann, Pastor, and
  Schaal]{ijspeert2013dynamical}
Ijspeert, A.~J., Nakanishi, J., Hoffmann, H., Pastor, P., and Schaal, S.
\newblock Dynamical movement primitives: learning attractor models for motor
  behaviors.
\newblock \emph{Neural computation}, 25\penalty0 (2):\penalty0 328--373, 2013.

\bibitem[Kim et~al.(2019)Kim, Ahn, and Bengio]{NIPS2019_9332}
Kim, T., Ahn, S., and Bengio, Y.
\newblock Variational temporal abstraction.
\newblock In \emph{Advances in Neural Information Processing Systems 32}, pp.\
  11566--11575. Curran Associates, Inc., 2019.
\newblock URL
  \url{http://papers.nips.cc/paper/9332-variational-temporal-abstraction.pdf}.

\bibitem[Kingma \& Ba(2014)Kingma and Ba]{kingma2014adam}
Kingma, D.~P. and Ba, J.
\newblock Adam: A method for stochastic optimization.
\newblock \emph{arXiv preprint arXiv:1412.6980}, 2014.

\bibitem[Kingma \& Welling(2013)Kingma and Welling]{kingma2013autoencoding}
Kingma, D.~P. and Welling, M.
\newblock Auto-encoding variational bayes, 2013.
\newblock URL \url{http://arxiv.org/abs/1312.6114}.
\newblock cite arxiv:1312.6114.

\bibitem[Kipf et~al.(2019)Kipf, Li, Dai, Zambaldi, Sanchez-Gonzalez,
  Grefenstette, Kohli, and Battaglia]{kipf2019compile}
Kipf, T., Li, Y., Dai, H., Zambaldi, V., Sanchez-Gonzalez, A., Grefenstette,
  E., Kohli, P., and Battaglia, P.
\newblock Compile: Compositional imitation learning and execution.
\newblock In \emph{ICML}, 2019.

\bibitem[Kober \& Peters(2009)Kober and Peters]{kober2009learning}
Kober, J. and Peters, J.
\newblock Learning motor primitives for robotics.
\newblock In \emph{ICRA}, 2009.

\bibitem[Konidaris \& Barto(2009)Konidaris and Barto]{konidaris2009skill}
Konidaris, G. and Barto, A.
\newblock Skill chaining: Skill discovery in continuous domains.
\newblock In \emph{the Multidisciplinary Symposium on Reinforcement Learning,
  Montreal, Canada}, 2009.

\bibitem[Konidaris et~al.(2012)Konidaris, Kuindersma, Grupen, and
  Barto]{konidaris2012robot}
Konidaris, G., Kuindersma, S., Grupen, R., and Barto, A.
\newblock Robot learning from demonstration by constructing skill trees.
\newblock \emph{IJRR}, 2012.

\bibitem[Kramer(1998)]{Kramer1998DirectedIF}
Kramer, G.
\newblock Directed information for channels with feedback.
\newblock 1998.

\bibitem[Krishnan et~al.(2017)Krishnan, Fox, Stoica, and
  Goldberg]{krishnan2017ddco}
Krishnan, S., Fox, R., Stoica, I., and Goldberg, K.
\newblock Ddco: Discovery of deep continuous options for robot learning from
  demonstrations.
\newblock \emph{arXiv preprint arXiv:1710.05421}, 2017.

\bibitem[Krishnan et~al.(2018)Krishnan, Garg, Patil, Lea, Hager, Abbeel, and
  Goldberg]{krishnan2018transition}
Krishnan, S., Garg, A., Patil, S., Lea, C., Hager, G., Abbeel, P., and
  Goldberg, K.
\newblock Transition state clustering: Unsupervised surgical trajectory
  segmentation for robot learning.
\newblock In \emph{RR}. 2018.

\bibitem[Kulkarni et~al.(2016)Kulkarni, Narasimhan, Saeedi, and
  Tenenbaum]{hrl_tejas}
Kulkarni, T.~D., Narasimhan, K.~R., Saeedi, A., and Tenenbaum, J.~B.
\newblock Hierarchical deep reinforcement learning: Integrating temporal
  abstraction and intrinsic motivation.
\newblock In \emph{Proceedings of the 30th International Conference on Neural
  Information Processing Systems}, NIPS’16, pp.\  3682–3690, Red Hook, NY,
  USA, 2016. Curran Associates Inc.
\newblock ISBN 9781510838819.

\bibitem[Leslie Pack~Kaelbling(2017)]{tlpkICRA17}
Leslie Pack~Kaelbling, T. L.-P.
\newblock Learning composable models of parameterized skills.
\newblock In \emph{IEEE Conference on Robotics and Automation (ICRA)}, 2017.
\newblock URL \url{http://lis.csail.mit.edu/pubs/lpk/ICRA17.pdf}.

\bibitem[Lillicrap et~al.(2015)Lillicrap, Hunt, Pritzel, Heess, Erez, Tassa,
  Silver, and Wierstra]{lillicrap2015continuous}
Lillicrap, T.~P., Hunt, J.~J., Pritzel, A., Heess, N., Erez, T., Tassa, Y.,
  Silver, D., and Wierstra, D.
\newblock Continuous control with deep reinforcement learning.
\newblock \emph{arXiv preprint arXiv:1509.02971}, 2015.

\bibitem[Lioutikov et~al.(2017)Lioutikov, Neumann, Maeda, and
  Peters]{doi:10.1177/0278364917713116}
Lioutikov, R., Neumann, G., Maeda, G., and Peters, J.
\newblock Learning movement primitive libraries through probabilistic
  segmentation.
\newblock \emph{The International Journal of Robotics Research}, 36\penalty0
  (8):\penalty0 879--894, 2017.
\newblock \doi{10.1177/0278364917713116}.
\newblock URL \url{https://doi.org/10.1177/0278364917713116}.

\bibitem[Lioutikov et~al.(2020)Lioutikov, Maeda, Veiga, Kersting, and
  Peters]{rudolf2}
Lioutikov, R., Maeda, G., Veiga, F., Kersting, K., and Peters, J.
\newblock Learning attribute grammars for movement primitive sequencing.
\newblock \emph{The International Journal of Robotics Research}, 39\penalty0
  (1):\penalty0 21--38, 2020.
\newblock \doi{10.1177/0278364919868279}.
\newblock URL \url{https://doi.org/10.1177/0278364919868279}.

\bibitem[Mandlekar et~al.(2018)Mandlekar, Zhu, Garg, Booher, Spero, Tung, Gao,
  Emmons, Gupta, Orbay, Savarese, and Fei-Fei]{mandlekar2018roboturk}
Mandlekar, A., Zhu, Y., Garg, A., Booher, J., Spero, M., Tung, A., Gao, J.,
  Emmons, J., Gupta, A., Orbay, E., Savarese, S., and Fei-Fei, L.
\newblock Roboturk: A crowdsourcing platform for robotic skill learning through
  imitation.
\newblock In \emph{Conference on Robot Learning}, 2018.

\bibitem[Meier et~al.(2011)Meier, Theodorou, Stulp, and
  Schaal]{meier2011movement}
Meier, F., Theodorou, E., Stulp, F., and Schaal, S.
\newblock Movement segmentation using a primitive library.
\newblock In \emph{2011 IEEE/RSJ International Conference on Intelligent Robots
  and Systems}, 2011.

\bibitem[{Muelling} et~al.(2010){Muelling}, {Kober}, and {Peters}]{5686298}
{Muelling}, K., {Kober}, J., and {Peters}, J.
\newblock Learning table tennis with a mixture of motor primitives.
\newblock In \emph{2010 10th IEEE-RAS International Conference on Humanoid
  Robots}, pp.\  411--416, Dec 2010.
\newblock \doi{10.1109/ICHR.2010.5686298}.

\bibitem[Murali et~al.(2016)Murali, Garg, Krishnan, Pokorny, Abbeel, Darrell,
  and Goldberg]{murali2016tsc}
Murali, A., Garg, A., Krishnan, S., Pokorny, F.~T., Abbeel, P., Darrell, T.,
  and Goldberg, K.
\newblock Tsc-dl: Unsupervised trajectory segmentation of multi-modal surgical
  demonstrations with deep learning.
\newblock In \emph{ICRA}, 2016.

\bibitem[Mülling et~al.(2013)Mülling, Kober, Kroemer, and
  Peters]{doi:10.1177/0278364912472380}
Mülling, K., Kober, J., Kroemer, O., and Peters, J.
\newblock Learning to select and generalize striking movements in robot table
  tennis.
\newblock \emph{The International Journal of Robotics Research}, 32\penalty0
  (3):\penalty0 263--279, 2013.
\newblock \doi{10.1177/0278364912472380}.
\newblock URL \url{https://doi.org/10.1177/0278364912472380}.

\bibitem[Neumann et~al.(2014)Neumann, Daniel, Paraschos, Kupcsik, and
  Peters]{neumann2014learning}
Neumann, G., Daniel, C., Paraschos, A., Kupcsik, A., and Peters, J.
\newblock Learning modular policies for robotics.
\newblock \emph{Frontiers in computational neuroscience}, 8:\penalty0 62, 2014.

\bibitem[Niekum et~al.(2012)Niekum, Osentoski, Konidaris, and
  Barto]{niekum2012learning}
Niekum, S., Osentoski, S., Konidaris, G., and Barto, A.~G.
\newblock Learning and generalization of complex tasks from unstructured
  demonstrations.
\newblock IEEE, 2012.

\bibitem[Peters et~al.(2013)Peters, Kober, M{\"u}lling, Kr{\"a}mer, and
  Neumann]{peters2013towards}
Peters, J., Kober, J., M{\"u}lling, K., Kr{\"a}mer, O., and Neumann, G.
\newblock Towards robot skill learning: From simple skills to table tennis.
\newblock In \emph{Joint European Conference on Machine Learning and Knowledge
  Discovery in Databases}, pp.\  627--631. Springer, 2013.

\bibitem[Schaal(1997)]{Schaal_ANIPS_1997}
Schaal, S.
\newblock Learning from demonstration.
\newblock In \emph{Advances in Neural Information Processing Systems 9}, pp.\
  1040--1046, Cambridge, MA, 1997. MIT Press.
\newblock URL \url{http://www-clmc.usc.edu/publications/S/schaal-NIPS1997.pdf}.
\newblock clmc.

\bibitem[Shankar et~al.(2020)Shankar, Tulsiani, Pinto, and
  Gupta]{shankar2020discovering}
Shankar, T., Tulsiani, S., Pinto, L., and Gupta, A.
\newblock Discovering motor programs by recomposing demonstrations.
\newblock In \emph{International Conference on Learning Representations}, 2020.
\newblock URL \url{https://openreview.net/forum?id=rkgHY0NYwr}.

\bibitem[Sharma et~al.(2019)Sharma, Sharma, Rhinehart, and
  Kitani]{DBLP:conf/iclr/SharmaSRK19}
Sharma, M., Sharma, A., Rhinehart, N., and Kitani, K.~M.
\newblock Directed-info {GAIL:} learning hierarchical policies from unsegmented
  demonstrations using directed information.
\newblock In \emph{7th International Conference on Learning Representations,
  {ICLR} 2019, New Orleans, LA, USA, May 6-9, 2019}, 2019.
\newblock URL \url{https://openreview.net/forum?id=BJeWUs05KQ}.

\bibitem[Sharma et~al.(2018)Sharma, Mohan, Pinto, and
  Gupta]{pmlr-v87-sharma18a}
Sharma, P., Mohan, L., Pinto, L., and Gupta, A.
\newblock Multiple interactions made easy (mime): Large scale demonstrations
  data for imitation.
\newblock In \emph{CoRL}, 2018.

\bibitem[Shiarlis et~al.(2018)Shiarlis, Wulfmeier, Salter, Whiteson, and
  Posner]{shiarlis2018taco}
Shiarlis, K., Wulfmeier, M., Salter, S., Whiteson, S., and Posner, I.
\newblock Taco: Learning task decomposition via temporal alignment for control.
\newblock \emph{arXiv preprint arXiv:1803.01840}, 2018.

\bibitem[Smith et~al.(2018)Smith, Hoof, and Pineau]{smith2018inference}
Smith, M., Hoof, H., and Pineau, J.
\newblock An inference-based policy gradient method for learning options.
\newblock In \emph{ICML}, 2018.

\bibitem[Sutton et~al.(1999)Sutton, Precup, and Singh]{sutton1999between}
Sutton, R.~S., Precup, D., and Singh, S.
\newblock Between mdps and semi-mdps: A framework for temporal abstraction in
  reinforcement learning.
\newblock \emph{Artificial intelligence}, 1999.

\bibitem[van~der Maaten \& Hinton(2008)van~der Maaten and
  Hinton]{maaten2008visualizing}
van~der Maaten, L. and Hinton, G.
\newblock Visualizing high-dimensional data using t-sne.
\newblock 2008.

\bibitem[Williams(1992)]{williams1992simple}
Williams, R.~J.
\newblock Simple statistical gradient-following algorithms for connectionist
  reinforcement learning.
\newblock \emph{Machine learning}, 1992.

\bibitem[Xu et~al.(2018)Xu, Nair, Zhu, Gao, Garg, Fei-Fei, and
  Savarese]{xu2018neural}
Xu, D., Nair, S., Zhu, Y., Gao, J., Garg, A., Fei-Fei, L., and Savarese, S.
\newblock Neural task programming: Learning to generalize across hierarchical
  tasks.
\newblock In \emph{ICRA}, 2018.

\bibitem[{Ziebart} et~al.(2013){Ziebart}, {Bagnell}, and {Dey}]{6479340}
{Ziebart}, B.~D., {Bagnell}, J.~A., and {Dey}, A.~K.
\newblock The principle of maximum causal entropy for estimating interacting
  processes.
\newblock \emph{IEEE Transactions on Information Theory}, 59\penalty0
  (4):\penalty0 1966--1980, April 2013.
\newblock ISSN 1557-9654.
\newblock \doi{10.1109/TIT.2012.2234824}.

\end{thebibliography}
\bibliographystyle{icml2019}

\newpage
\onecolumn
\section{Appendix}
We provide additional details and insights into our approach below.

\subsection{Derivation of Temporal Variational Inference:}
While our main paper presents the specific temporal variational inference that we use to learn policies, the notion of temporal variational inference is more generally applicable to sequential data with hidden states (such as in HMMs). 

We present a detailed derivation of this general temporal variational inference objective, and contrast it with standard variational inference below. We then provide a detailed derivation of the gradient of this objective in our case, explaining how dependencies on system dynamics may be factored out. 


We begin by considering observed sequential data $x = \{ x_t \}_{t=1}^t$ (in our case, this corresponds to state-action tuples, i.e. $\{ x_t = (s_t,a_t) \}_{t=1}^{T}$), interacting with a sequence of unobserved latent variables $y = \{ y_t \}_{t=1}^T$ (in our case, $y_t$ is simply $\zeta_t$).
For these sequences of data, the true likelihood of observed data $\mathcal{L} = \mathbb{E}_{x \sim p(x)} \big[ \log p(x) \big]$ is lower bounded by $J$, where:
\begin{align*}
     J &= \mathbb{E}_{x \sim p(x)} \big[ \log p(x) \big] - D_{KL} \big[ q(y|x) || p(y|x) \big]
\end{align*} 
where $q(y|x)$ is a variational approximation to the true conditional $p(y|x)$, and the lower bounded due to the non-negativity of KL divergence. Expanding the KL divergence term, and using the fact that $p(x,y) = p(y|x) p(x)$, we have:
\begin{align*}
    J &= \mathbb{E}_{x \sim p(x)} \big[ \log p(x) \big] - \mathbb{E}_{x \sim p(x), y \sim q(y|x)} \Big[ \log \frac{q(y|x)}{p(y|x)} \Big] \\
    J &= \mathbb{E}_{x \sim p(x), y \sim q(y|x)} \Big[ \log p(x,y) - \log q(y|x) \Big] 
\end{align*} 
Standard variational inference then decomposes the joint likelihood $p(x,y)$ into a ``decoder'' $p(x|y)$ and a prior $p(y)$:
\begin{align*}
    J_{\rm StandardVI} &= \mathbb{E}_{x \sim p(x), y \sim q(y|x)} \Big[ \log p(x|y) + \log p(y) - \log q(y|x) \Big]
\end{align*}
Given the sequential nature of $x$ and $y$, inferring the conditional $p(x|y)$ does not provide a useful insight into how the \textit{sequence} of $x$ and $y$ will evolve, and is often difficult to learn.
In contrast, our temporal variational inference makes use of the following decomposition of joint likelihood $p(x,y)$: 
\begin{align*}
    p(x,y) &= \underbrace{\prod_{t=1}^T p(x_t|x_{1:t-1},y_{1:t-1})}_{\text{$= p(x||y)$}} \underbrace{\prod_{t=1}^T p(y_t|x_{1:t},y_{1:t-1})}_{\text{$= p(y||x)$}}
\end{align*}
$p(x||y)$ and $p(y||x)$ are \textit{causally conditioned distributions}, i.e. they only depend on information available until the current timestep. $p(x||y)$ and $p(y||x)$ are formally defined as $\prod_{t=1}^T p(x_t|x_{1:t-1},y_{1:t-1})$ and $\prod_{t=1}^T p(y_t|x_{1:t},y_{1:t-1})$ respectively \cite{6479340}. We can plug this decomposition into the objective $J$  to give us the temporal variational inference objective:
\begin{align*}
    J_{\rm Temporal VI} &= \mathbb{E}_{x \sim p(x), y \sim q(y|x)} \Big[ \log p(x||y) + \log p(y||x) - \log q(y|x) \Big]\\
    &= \mathbb{E}_{x \sim p(x), y \sim q(y|x)} \Big[ \log \prod_{t=1}^T p(x_t|x_{1:t-1}, y_{1:t-1}) + \log \prod_{t=1}^T  p(y_t|x_{1:t},y_{1:t-1}) - \log q(y|x) \Big] \\
    &= \mathbb{E}_{x \sim p(x), y \sim q(y|x)} \Big[ \sum_{t=1}^T \log p(x_t|x_{1:t-1}, y_{1:t-1}) + \sum_{t=1}^T \log p(y_t|x_{1:t},y_{1:t-1}) - \log q(y|x) \Big]
\end{align*}


\subsection{Derivation of Gradient Update:}
Now that we have a better understanding of the origin of the temporal variational inference objective, we present a derivation of the gradient to this objective in our case. In our option learning setting, the \tvi  objective is:
\begin{align*}
    J_{\rm TemporalVI} &= \mathbb{E}_{\tau \sim \mathcal{D}, \zeta \sim q(\zeta|\tau)} \Big[ \sum_{t=1}^T \big\{ \log \eta(\zeta_t | s_{1:t}, a_{1:t-1}, \zeta_{1:t-1}) + \log \pi(a_t | s_{1:t}, a_{1:t-1}, \zeta_{1:t}) \\  
    &+ \log p(s_{t+1} | s_t, a_t) \big\} + \log p(s_1) - \log q(\zeta|\tau) \Big]
\end{align*}
Assuming distributions $\pi, \eta$, and $q$ are parameterized by $\theta, \phi$, and $\omega$ respectively, the gradient of this objective is:
\begin{align*}
    \nabla_{\theta, \phi, \omega} J_{\rm TemporalVI} &= \nabla_{\theta, \phi, \omega} \ \mathbb{E}_{\tau \sim \mathcal{D}, \zeta \sim q(\zeta|\tau)} \Big[ \sum_{t=1}^T \big\{ \log \eta(\zeta_t | s_{1:t}, a_{1:t-1}, \zeta_{1:t-1}) + \log \pi(a_t | s_{1:t}, a_{1:t-1}, \zeta_{1:t}) \\  
    &+ \log p(s_{t+1} | s_t, a_t) \big\} + \log p(s_1) - \log q(\zeta|\tau) \Big]
\end{align*}
Note that the system dynamics $p(s_{t+1}|s_t,a_t)$ and the initial state distribution $p(s_1)$ are both independent of the parameterization of networks $\theta, \phi$, and $\omega$. The gradient of the objective with respect to $\theta, \phi$, and $\omega$ can be separated into additive terms that depend on the dynamics and initial state distributions, and terms that depend on networks $\pi, \eta$, and $q$. The expectation of the dynamics and initial state distribution terms are constant, and their gradient hence vanishes:
\begin{align*}
    \nabla_{\theta, \phi, \omega} J_{\rm TemporalVI} &= \nabla_{\theta, \phi, \omega} \  \underbrace{\mathbb{E}_{\tau \sim \mathcal{D}, \zeta \sim q(\zeta|\tau)} \Big[ \sum_{t=1}^T \log p(s_{t+1} | s_t, a_t) + \log p(s_1) \Big]}_{\text{Constant}} = 0
\end{align*}
The gradient update of temporal variational inference hence doesn't depend on the dynamics and the initial state distribution, leading to the following gradient update, as presented in the main paper in equation 3:
\begin{align*}
    \nabla_{\theta, \phi, \omega} J &= \nabla_{\theta, \phi, \omega} \mathbb{E}_{\tau \sim \mathcal{D}, \zeta \sim q(\zeta|\tau)} \Big[ \sum_t \log \pi(a_t | s_{1:t}, a_{1:t-1}, \zeta_{1:t}) + \sum_t \log \eta(b_t, z_t | s_{1:t}, a_{1:t-1}, \zeta_{1:t-1}) - \log q(\zeta|\tau) \Big] 
\end{align*}

\subsection{Implementation Details:}
We make note of several implementation details below, such as network architectures and hyperparameter settings. 

\subsubsection{Network Architecture:}
We describe the specific architectures of each of the networks $q, \pi$, and $\eta$ in our approach. The base architecture for each of these three networks is an $8$ layer LSTM with $128$ hidden units in each layer. We found that an $8$ layer LSTM was sufficiently expressive for represent the distributions $q, \pi$, and $\eta$. 

The space of predictions for $q$ and $\eta$ are the binary termination variables $b_t$ at every timestep $t$, and the continuous parameterization of options $z$. 
$q$ and $\eta$ are thus implemented with two heads on top of the final LSTM layer. 

The first head predicts termination probabilities (from which termination variables $b_t$ are sampled), and consists of a linear layer of output size $2$ followed by a softmax layer to predict probabilities. 

The second head of the network predicts the latent $z$ parameterization at each timestep. It consists of two separate linear layers above the final LSTM layer, that predict the mean and variance of a Gaussian distribution respectively. The mean predictions do not use an activation layer. The variance predictions employ a SoftPlus activation function to predict positive variances. The dimensionality of latent $z$'s is $64$ across all three datasets and across all networks. 

The prediction space for $\pi$ is the continuous low-level actions (i.e. joint velocities). Similar to the $z$ prediction, this is implemented by
with 2 separate linear layers to predict the mean and variances of a Gaussian distribution, from which actions are drawn. As above, variances use a SoftPlus activation, while mean predictions are done without an activation.
The dimensions of the action space are $16$ for the MIME dataset, $8$ for the Roboturk dataset, and $66$ for the Mocap dataset. 

\subsubsection{Hyperparameters:}
We provide a list of the hyperparameters and their values, and other training details used in our training procedure. 
\begin{enumerate}
    \item Optimizer: We use the Adam optimizer \cite{kingma2014adam} to train all networks in our model. 
    \item Learning Rate: We use a learning rate of $10^{-4}$ for our optimizer, as is standard. 
    \item Epsilon: The exploration parameter $\epsilon$ is used in our training procedure to both scale perturbation of latent $z$'s sampled from our model, as well as to explore different latent $b$'s in an epsilon-greedy fashion. We use an initial $\epsilon$ value of $0.3$, and linearly decay this value to $0.05$ over $30$ epochs of training, and found this works well. 
    We considered a range of $0.1-0.3$ for the initial value of epsilon, and a range of $0.05-0.1$ for the final epsilon. 
    \item Ornstein Uhlenbeck Noise Parameters: Our DDPG implementation for the RL training uses the Ornstein Uhlenbeck noise process, with parameters identical to those used in the DDPG paper \cite{lillicrap2015continuous}.
    \item Loss Weights: In practice, the various terms in our objective are reweighted prior to gradient computation, to facilitate learning the desired behaviors and to prevent particular terms from dominating others. 
    \begin{enumerate}
        \item Option likelihood weight: During initial phases of training, we reweight the option likelihood term $\sum_t \log \eta(\zeta_t | s_{1:t}, a_{1:t-1}, \zeta_{1:t-1})$ in our gradient update by a factor of $0.01$, to prevent the variational network from getting inconsistent gradients from the randomly initialized $\eta$. Once the variational policy has been trained sufficiently, we set the weight of this option likelihood to $1$.
        \item KL Divergence weight: We reweight the KL divergence term, as done in the $\beta$-VAE paper \cite{higgins2017beta}, by a factor of $0.01$. 
    \end{enumerate}
\end{enumerate}

\subsubsection{RL Details:}
For our reinforcement learning experiments, we use variants of the following Robosuite \cite{mandlekar2018roboturk} environments to evaluate our approach: 
\begin{itemize}
    \item SawyerPickPlace - An environment where a sawyer robot grasps and places objects in specific positions in respective bins. We use $4$ variants of this task, SawyerPickPlaceBread, SawyerPickPlaceCereal, SawyerPickPlaceCan, SawyerPickPlaceMilk, where the objective is to place the corresponding object into the correct bin. 
    
    \item SawyerNutAssembly - An environment where a sawyer robot must pick a nut up and place it around an appropriately shaped peg. We use $2$ variants of this task, SawyerNutAssemblySquare and SawyerNutAssemblyRound, where the shape of the nut and peg are varied. 
    
\end{itemize}
The $3$ baseline algorithms specified in the main paper and our approach all share the same policy architectures (i.e., an $8$ layer LSTM with $128$ hidden units) for both low-level policies (in all baselines and our approach) and high-level policies (our approach and the hierarchical RL baseline).

For these pick-place and nut-assembly tasks, the information available to the policies are the sequence of  joint states of the robot, previous joint velocities executed, the \texttt{robot-state} provided by Robosuite (consists of sin and cos of the joint angles, gripper positions, etc.), and the \texttt{object-state} provided by Robosuite (consisting of absolute positions of the target objects, object positions relative to the robot end-effector, etc.). The output space for the policies is always the joint velocities (including the gripper).

\subsubsection{Dataset details:}
Regarding the CMU Mocap dataset, the data used in this project was obtained from mocap.cs.cmu.edu, the database was created with funding from NSF EIA-0196217.

\end{document}